\documentclass{article}
\usepackage{ijcai25}

\usepackage{times}
\usepackage{soul}
\usepackage{url}
\usepackage[hidelinks]{hyperref}
\usepackage[utf8]{inputenc}
\usepackage[small]{caption}
\usepackage{graphicx}
\usepackage{amsmath}
\usepackage{amsthm}
\usepackage{booktabs}
\usepackage{algorithm}
\usepackage[switch]{lineno}
\usepackage{bbding}

\usepackage{amssymb}
\usepackage{multirow}
\usepackage{colortbl}
\usepackage{enumitem}
\usepackage{algpseudocode}
\usepackage{subcaption}
\usepackage[dvipsnames]{xcolor}
\usepackage{bm}

\title{Self-Classification Enhancement and Correction \\ for Weakly Supervised Object Detection}

\author{
Yufei Yin$^1$
\and
Lechao Cheng$^2$\thanks{Corresponding authors: Lechao Cheng and Zhou Yu}\and
Wengang Zhou$^{3}$\and
Jiajun Deng$^{4}$\and
Zhou Yu$^{1}$\footnotemark[1]\And
Houqiang Li$^{3}$\\
\affiliations
$^1$School of Computer Science, Hangzhou Dianzi University\\
$^2$School of Computer Science and Information Engineering, Hefei University of Technology\\
$^3$EEIS Department, University of Science and Technology of China\\
$^4$Australian Institute for Machine Learning, University of Adelaide\\
\emails
yinyf@hdu.edu.cn,
chenglc@hfut.edu.cn,
zhwg@ustc.edu.cn, \\
jiajun.deng@adelaide.edu.au,
yuz@hdu.edu.cn,
lihq@ustc.edu.cn
}

\begin{document}

\maketitle

\begin{abstract}
    In recent years, weakly supervised object detection (WSOD) has attracted much attention due to its low labeling cost. The success of recent WSOD models is often ascribed to the two-stage multi-class classification (MCC) task, \textit{i.e.}, multiple instance learning and online classification refinement.  Despite  achieving non-trivial progresses, these methods overlook potential classification ambiguities between these two  MCC tasks and fail to leverage their unique strengths.  In this work, we introduce a novel WSOD framework to ameliorate these two issues. For one thing, we propose a self-classification enhancement module that integrates intra-class binary classification (ICBC) to bridge the gap between the two distinct MCC tasks. The ICBC task enhances the network’s discrimination between positive and mis-located samples in a class-wise manner and forges a mutually reinforcing relationship with the MCC task. For another, we propose a self-classification correction algorithm during inference, which combines the results of both MCC tasks to effectively reduce the mis-classified predictions. Extensive experiments on the prevalent VOC 2007 \& 2012 datasets demonstrate the superior performance of our framework.
\end{abstract}

\section{Introduction}
\label{sec:intro}
Object detection aims to localize objects of interest and classify them, which is a fundamental task in the field of computer vision.
The recent decade has witnessed rapid progress \cite{FastR-CNN,FasterR-CNN,SSD,YOLO} in various object detection scenarios~\cite{nie2023adapting,wang2023biased,jiao2024instance,wang2024marvelovd}, benefiting from the development of convolutional neural networks (CNN). In spite of the remarkable advances, current fine-grained instance-level annotations are labor-intensive and time-consuming to obtain. This paper focuses on the domain of weakly supervised object detection (WSOD)~\cite{su2022re}, which requires only image-level annotations, \textit{i.e.}, existing object categories in a given image, to achieve the object detection task.

\begin{figure}[t!]
	\centering
	\includegraphics[scale = 0.42]{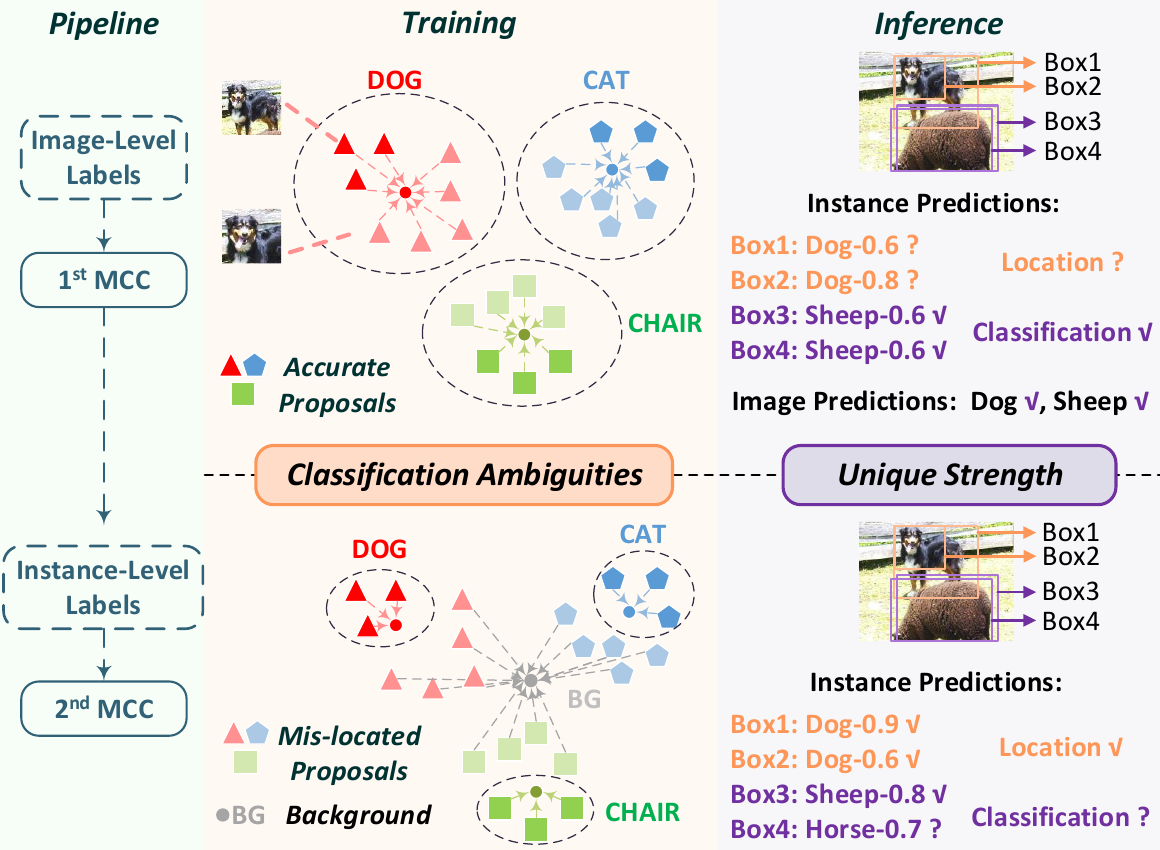} \\
	\caption{Comparison between the two distinct MCC tasks.} 
	\label{img:intro}
\end{figure}

Recent WSOD approaches~\cite{OICR,ts2c,MIST,CASD} generally convert WSOD into a two-stage multi-class classification (MCC) pipeline. In the first stage, a multiple instance detection network (MIDN)~\cite{WSDDN} is constructed, leveraging multiple instance learning to introduce competition both among object classes and proposals (denoted as 1$^{st}$ MCC). This process effectively identifies candidate regions that contain significant class-specific patterns. However, MIDN often suffers from partial-located issues, wherein high scores are assigned to the detections localizing only the most discriminative parts. To overcome this challenge, in the second stage, cascaded online multi-class classifiers \cite{OICR,SLV} are integrated to refine the classification scores of MIDN, and assorted strategies \cite{WSOD2,MIST} are designed to generate pseudo instance-level labels for training these classifiers (denoted as 2$^{nd}$ MCC). 
Despite the promising results achieved by these methods, as shown in Figure~\ref{img:intro}, they overlook the potential classification ambiguities and the unique strengths of the two distinct multi-class classification tasks across the two stages:

\begin{enumerate}[label=(\roman*)]
	\item During the 1$^{st}$ MCC task, some mis-located proposals, especially those only containing discriminative parts of an object, are classified as the corresponding object class and leveraged to generate its class-specific features. However, after multiple refinements in the 2$^{nd}$ MCC task, these proposals will be classified as the background class, whose features are instead pushed toward those of mis-located proposals from other classes. These ambiguities compromise the quality of the class-specific features produced by the detector.
    \item  These two MCC tasks are guided by image-level and pseudo-instance-level labels, respectively. As a result, the first MCC task excels at identifying the classes present in the image, while the second one focuses on more accurate instance-level location. However, previous methods rely solely on the second MCC task during inference, overlooking the classification benefits provided by the first task.
\end{enumerate}

In this paper, We present a novel \underline{\textbf{s}}elf-\underline{\textbf{c}}lassification \underline{\textbf{e}}nhancement and \underline{\textbf{c}}orrection (SCEC) framework to overcome these two limitations.
To alleviate the \textit{classification ambiguities}, we introduce a self-classification enhancement module during the second stage, which integrates an extra intra-class binary classification (ICBC) task to bridge the gap between the two distinct MCC tasks. ICBC task aims to enhance the network's discrimination between positive and mis-located samples in a class-specific manner, 
rather than directly grouping them together with background samples into a single `background' class, as done in the 2$^{nd}$ MCC.
To sufficiently optimize the ICBC classifiers, we generate various types of mis-located samples based on the 2$^{nd}$ MCC results.
Furthermore, the ICBC results are utilized in reverse to refine the pseudo labels for the 2$^{nd}$ MCC task, allowing the two tasks to complement each other.
To harness the \textit{unique strengths} of the two distinct MCC tasks, we introduce a self-classification correction algorithm, which leverages the 1$^{st}$ MCC results to rectify mis-classifications in the detections produced by the second one.

Our primary contributions are summarized as follows:
\begin{itemize} [noitemsep,nolistsep]
    \item We propose a self-classification enhancement module that incorporates both the base multi-class classification and an intra-class binary classification to alleviate the classification ambiguities. These two tasks are carried out in a mutually reinforcing manner. 
    \item We introduce a self-classification correction algorithm to alleviate the mis-classification problem of detections, leveraging the image-level classification strength of the first multi-class classification results during inference.
    \item Extensive experiments on the prevalent PASCAL VOC 2007 and 2012 datasets demonstrate the superior performance of our framework.

\end{itemize}

\section{Related Work}
\textbf{Weakly Supervised Object Detection.}  Weakly Supervised Object Detection (WSOD) has been widely studied in recent years. The pioneering work WSDDN \cite{WSDDN} first integrates multiple instance learning into the CNN architecture by designing a two-stream network, \textit{i.e.} classification branch and detection branch. By combining the results from these two branches, WSDDN converts the WSOD task into a multi-class classification problem for proposals. However, such a solution often struggles to generate accurate detections. To alleviate this problem, OICR \cite{OICR} proposes a two-stage pipeline where WSDDN is utilized as a basic detector, and its results are utilized to generate pseudo seed boxes for training several subsequent online instance classifiers for further refinement. Most recent WSOD approaches are developed based on this pipeline. 
Some methods improve the detection capability of the basic detector, \textit{e.g.}, adding extra supplement modules for more complete detections \cite{C-midn,IMCFB,yin2022fi}, and enhancing the generated image or proposal feature \cite{MIST,CASD}. 
Some other methods design various strategies to improve the quality of pseudo seed boxes, \textit{e.g.}, constructing spatial graphs \cite{PCL} or appearance graphs \cite{OIM} for top-scoring proposals, bringing top-down objectness \cite{WSOD2}, and applying spatial likelihood voting \cite{SLV}. Otherwise, \cite{Yang_2019_ICCV} add online regression branches to refine the initial proposals.

Different from them, our method extends the widely used online classifier into a self-classification enhancement module, which brings intra-class binary classification to enhance the network's discrimination between class-specific positive and mis-located samples, thus improving the detection capability of the network.

\section{Method}
\subsection{Overview}
\begin{figure*}[t!]
	\centering
	\includegraphics[scale = 0.8]{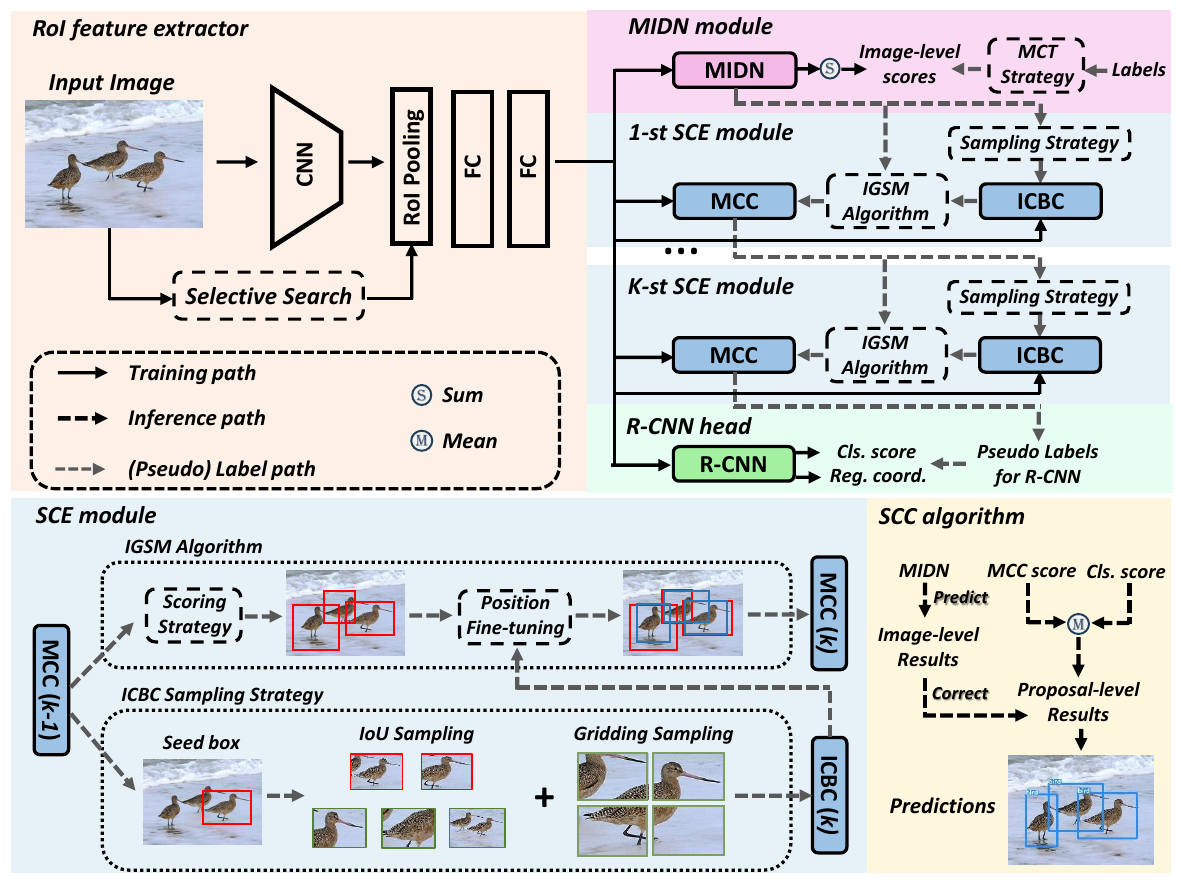} \\
	\vspace{-0.1in}
    	\caption{An overview of our self-classification enhancement and correction (SCEC) framework.} 
	\label{img:framework}
	\vspace{-0.1in}
\end{figure*}

The overview of the proposed model is illustrated in Figure~\ref{img:framework}.  First, an image and the generated region proposals are fed to the RoI feature extractor, \textit{i.e.}, CNN backbone and an RoI pooling layer followed by two FC layers, to obtain proposal feature vectors. Next, the feature factors are fed into MIDN module to produce instance-level scores, which are summed for the training of MIDN with image-level labels. 
Meanwhile, the feature factors are fed into several subsequent self-classification enhancement (SCE) module to obtain MCC and ICBC scores. For each SCE module, the ICBC branch is trained using MCC scores, while the pseudo labels of MCC branch are generated from ICBC Guided Seed Mining (IGSM) algorithm. 
After that, an R-CNN head is constructed to produce classification scores and regression coordinates. 
During inference, self-classification correction (SCC) algorithm is applied to generate prediction results.

\subsection{Basic WSOD Framework}
In the weakly supervised setting, distinguishing between positive and negative proposals directly during training becomes challenging, given the absence of instance-level labels.  To overcome this problem,  the pioneering work WSDDN \cite{WSDDN} adopts multiple instance learning into the CNN architecture to convert the WSOD task into the multi-class classification task for proposals. Following recent works \cite{OICR,MIST}, we apply WSDDN as our basic detector, referred as Multiple Instance Detector Network (MIDN).

Given an image $I$, its image-level labels $Y = [y_1, y_2, \cdots, y_C] \in \mathbb{R}^{C \times 1} $ is available according to the WSOD setting, where $y_c \in \{0, 1\}$ indicates the presence or absence of  class $c$. Its proposals $R = \left\{R_1, R_2, \cdots, R_N\right\}$ are pre-generated from Selective Search \cite{SS} before training. First, the proposal feature vectors are generated through a CNN backbone, an RoI pooling layer \cite{FastR-CNN}, and two FC layers. Next, these vectors are fed into two parallel branches, \textit{i.e.}, classification and detection branches, to obtain proposal scores. For each branch, a scoring matrix $x^\text{cls} (x^\text{det})  \in\mathbb{R}^{C \times |R|}$ is first obtained by an FC layer, where $|R|$ and $ C$ represents the number of proposals and categories, respectively. The two scoring matrices are then normalized by the softmax operation through orthogonal directions, \textit{i.e.} $\sigma(x^\text{cls})$ for category direction and $\sigma(x^\text{det})$ for proposal direction.
After that, the proposal scores are generated by the element-wise product of these two matrices: $x^\text{box} = \sigma(x^\text{cls}) \odot \sigma(x^\text{det})$. Finally, the image-level scores are obtained by aggregating over the proposal direction of $x^\text{box}$: $x^\text{img}_c = \sum_{i=1}^{|R|}x^\text{box}_{c, i}$. In this way, the image-level labels can be utilized for supervision through binary cross-entropy loss: $\mathcal{L}_{MIDN}=-\sum_{c=1}^C\left[y_clogx^\text{img}_c+\left(1-y_c\right)log\left(1-x^\text{img}_c\right)\right].$

Following OICR \cite{OICR}, the basic WSOD framework adds several online instance classification (OIC) branches after the basic detector to generate more accurate detections. Each branch contains an FC layer and a softmax operation and outputs a scoring matrix $x^\text{oic} \in\mathbb{R}^{(C+1) \times |R|}$. The top-scoring proposals from the $C$-th branch are utilized to generate pseudo labels $y^\text{oic}$ to train the $C+1$-th branch, through the cross-entropy loss: $\mathcal{L}_{OIC}=-\frac{1}{|R|}\sum_{i=1}^{|R|}\sum_{c=1}^{C+1}w_iy_{c,i}^\text{oic}logx_{c,i}^\text{oic}$. The loss weight $w_i$, which acts as a confidence score, is obtained from the score of the seed box which has the highest overlaps with $R_i$.
Additionally, following \cite{Yang_2019_ICCV,IMCFB}, we construct an R-CNN branch subsequently, which contains a classification sub-branch and a regression sub-branch. The weighted cross-entropy loss and smooth-L1 loss are applied to train these two sub-branches, respectively.

\subsection{Self-Classification Enhancement}
The basic WSOD framework refines the initial detection results by applying OIC branches with multi-class classification. However, as illustrated in Sec.~\ref{sec:intro}, this approach lead to classification ambiguities, where the features of mis-located samples from all classes are pushed together, despite that they are utilized to generate class-specific features during MIDN's training process.
Furthermore, such a solution will weaken the model's ability to distinguish the mis-located samples from their closed intra-class positive ones. To this end, we introduce Self-Classification Enhancement (SEC) module to tackle this problem. 

The SCE module comprises two parallel branches: one for the base multi-class classification and the other for an enhanced intra-class binary classification. These branches work harmoniously during online training, supplementing each other's strengths.
{\flushleft \bf Multi-Class Classification.} The multi-class classification (MCC) layer shares the same structure with the original OIC layer, containing an FC layer followed by a softmax.

{\flushleft \bf Intra-Class Binary Classification.} 
We incorporate the intra-class binary classification (ICBC) task to enhance the network's  discrimination between intra-class positive and mis-located samples. To maintain consistency with the MCC layer, we adopt a simple yet effective design to achieve the ICBC task. Specifically, given the proposal feature vector $f$, the ICBC  branch consists of an FC layer to generate score matrices and a sigmoid function for normalization:
\begin{equation}
\label{equ:icbc}
    x^{\text{icbc}}_{c,i} = \sigma(FC(f)), \quad x^{\text{icbc}} \in \mathbb{R}^{C \times |R|},
\end{equation}
where a higher $x^{\text{icbc}}_{c,i}$ indicates that the proposal $R_i$ is more likely to be a positive sample for class $c$, while a lower value suggests the opposite.

{\flushleft \bf Sampling strategy for ICBC task.} 
We adopt MCC results to select training samples $U$ for the ICBC task. Specifically, we first choose a set of top-scoring proposals as the pseudo seed boxes $S = \left\{S_1, S_2, \cdots, S_N\right\}$ according to MCC results. After that, we propose different strategies to select positive and mis-located samples based on these seed boxes. 

A straightforward way is to apply the \textit{IoU sampling strategy}. Specifically, for each proposal, we calculate its Intersection over Union (IoU) with all seed boxes, and take the maximum value $IoU_i$.
Next, we denote the positive samples as $R_\text{pos} = \left\{R_i| IoU_i  \ge \tau_h\right\}$ and the mis-located samples as $R_\text{neg} = \left\{R_i| \tau_l\le IoU_i  < \tau_h \right\}$, where $IoU_i$ is the overlaps between $R_i$ and its closest seed box $S_j$. Correspondingly, $R_i$ shares the same category $C_i$ with $S_j$.
Then, we generate the pseudo label $Y^{\text{icbc}}_i = [y^{\text{icbc}}_{1,i}, y^{\text{icbc}}_{2,i}, \cdots , y^{\text{icbc}}_{C,i}]$ of proposal $i$ according to the divisions:

\begin{equation}
\label{equ:p_label}
     y^{\text{icbc}}_{c,i}=\left\{
                \begin{aligned}
                1, & \quad R_i \in R_\text{pos} \ and \ C_i = c, \\
                0, & \quad else.
                \end{aligned}
        \right.
\end{equation}

However, the samples selected from IoU sampling strategy are insufficient for achieving the ICBC task. Given the limited number of training samples, we further apply a \textit{gridding sampling strategy} to generate additional mis-located samples for augmentation. Specifically, for each seed box with class $c$, we first apply box scaling by a scaling factor $\theta = 0.5$. In this way, the width and height of the
seed box are randomly sampled in $[(1 - \theta) w,(1 + \theta )w]$ and $[(1 - \theta )h,(1 + \theta )h]$, respectively. The center of the box remains unchanged.
Afterward, we generate an $n \times n$ grid on the scaled seed box. We treat each grid as a potential mis-located sample that may only contain a part of an object. It is worth noting that although some grids containing only background may inevitably be selected, their impact is minimal due to their limited quantity. These selected grids $G = \left\{G_1, G_2, \cdots, G_N\right\}$ are then fed to the RoI pooling layer to generate region features, which are subsequently passed to the ICBC layer. Their pseudo labels are assigned as follows: $y^{\text{icbc}}_{c,G_i}=0, G_i \in G$.

We denote all the selected samples as $U = R_\text{pos} \cup R_\text{neg} \cup G$. 
Considering the division of positive and mis-located samples is class-specific, we utilize sample weight to ensure these samples only participate in the losses of their corresponding categories:
\begin{equation}
\label{equ:p_weight}
     w^{\text{icbc}}_{c,i}=\left\{
                \begin{aligned}
                p_i, & \quad U_i \in R_\text{pos} \cup R_\text{neg} \ and \ C_i = c, \\
                q_i, & \quad U_i \in G \ and \ C_i = c, \\
                0, & \quad else,
                \end{aligned}
        \right.
\end{equation}
where $p_i$ is the confidence of $R_i$, obtained by the MCC score of its closest seed box $S_j$, and $q_i$ is set to 1.5. 
Finally, we adopt weighted binary cross-entropy loss for training ICBC layer:
\begin{equation}
\label{icbc_loss}
\begin{aligned}
& \mathcal{L}_{ICBC}=-\frac{1}{|U|}\sum_{i=1}^{|U|}\sum_{c=1}^{C}w^{\text{icbc}}_{c,i}BCE(x^{\text{icbc}}_{c,i}, y^{\text{icbc}}_{c, i}).
\end{aligned}
\end{equation}

\begin{figure}[t!]
	\centering
	\includegraphics[scale = 0.32]{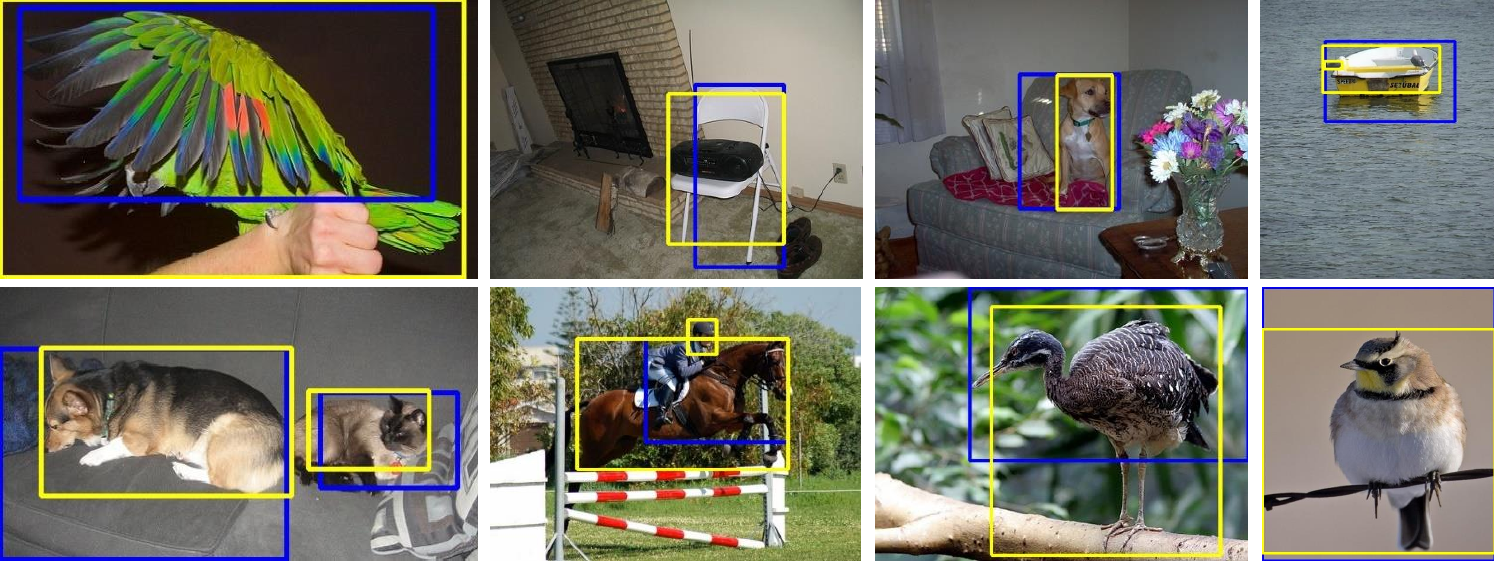} \\
	\caption{Comparison between the seed boxes selected from MCC scores (\textcolor{blue}{blue}) and ICBC scores (\textcolor{Goldenrod}{yellow}).}
	\label{img:icbc}
	\vspace{-0.15in}
\end{figure}

{\flushleft \bf ICBC Guided Seed Mining.} 
Seed box mining plays a significant role in training the MCC branch. Some methods pursue the accuracy of seed boxes by using top-scoring strategies \cite{OICR}, while some others focus on the recall of seed boxes by relaxing the top criteria and adopting non-maximum suppression (NMS) algorithm to remove redundant ones \cite{MIST}. Different from them, we adopt a soft scoring threshold to mine accurate seed boxes and utilize ICBC results for further fine-tuning.

Specifically, for each existing class $c$ ($y_c = 1$), we first find the proposal $R_k$ with the top score $x^{\text{mcc}}_{c,k}$. Here, we adopt the scores from the previous MCC branch following \cite{OICR}.  Next, instead of selecting top-$K$ proposals, we set a soft scoring threshold according to the top score in this class: $\tau_{score} = \alpha x^{\text{mcc}}_{c,k}$. We select the proposals whose MCC scores are higher than the threshold $\tau_{score}$. After that, we apply NMS algorithm on the selected proposals to remove redundant boxes, obtaining the base seed boxes $S_{base}$. 

Compared to the MCC, ICBC demonstrates superior proficiency in discerning the positive samples from their closed but mis-located ones in a class-specific manner. To this end, we apply ICBC results to fine-tune the obtained base seed boxes $S_{base}$. Briefly, for each seed box $S_i$ in $S_{base}$, we first obtain its surrounding proposals by setting an overlap threshold ($\tau_{sur}=0.5$). Next, among all the surrounding proposals, we select the one $S_{\hat{i}}$ that has the maximum ICBC score in the class of $S_i$, and add it to the seed boxes.  According to the previous comparison between ICBC and MCC, $S_{\hat{i}}$ can be regarded as a potential refinement of $S_i$ in location, which is shown in Figure \ref{img:icbc}. 

Finally, the fine-tuned seed boxes $S_{ft}$, along with the base seed boxes $S_{base}$, are utilized to train the MCC layer. We generate pseudo labels of all the proposals according to their max overlaps with seed boxes. The overlap thresholds are set the same with those in ICBC training, while the mis-located samples are labeled as $C+1$.
After that, we use these pseudo labels to train  MCC layer with weighted cross-entropy loss:
\begin{equation}
\label{mcc_loss}
\mathcal{L}_{MCC}=-\frac{1}{|R|}\sum_{i=1}^{|R|}\sum_{c=1}^{C+1}w^\text{mcc}_iy_{c,i}^\text{mcc}logx_{c,i}^\text{mcc},
\end{equation}
where $y_{c,i}^\text{mcc}$ and $x_{c,i}^\text{mcc}$ represent the pseudo labels and MCC score of proposal $R_i$ in class $c$, respectively. The loss weight $w^\text{mcc}_i$ is the score of the seed box which has the highest overlaps with $R_i$. 

Finally, we replace the original OIC branches with our SEC modules, and we train the network end-to-end by combining all the losses mentioned before:
\begin{equation}
\label{total_loss}
\mathcal{L}_{total}=\mathcal{L}_{MIDN} + \sum_{t=1}^{T}(\mathcal{L}_{MCC} + \gamma\mathcal{L}_{ICBC}) + \mathcal{L}_{R-CNN},
\end{equation}
where $\mathcal{L}_{R-CNN}$ is the loss for R-CNN branch, $T$ is the number of  online enhanced instance classification modules, and $\gamma$ keeps the balance between ${L}_{MCC}$ and ${L}_{ICBC}$.

\begin{figure}[t!]
	\centering
	\includegraphics[scale = 0.45]{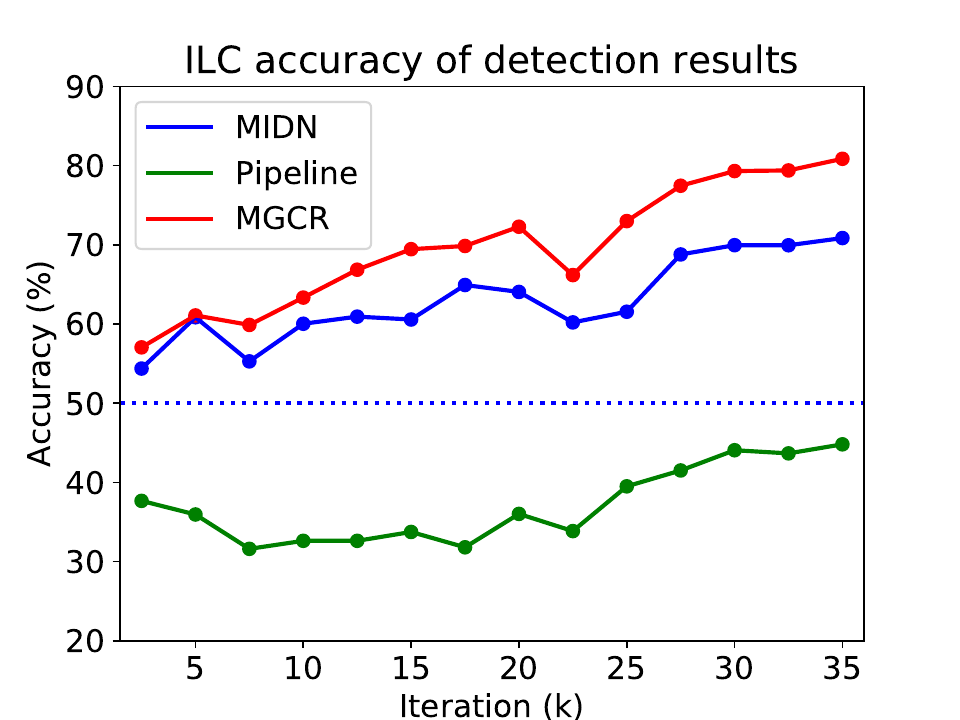} \\
    \vspace{-0.1in}
	\caption{Comparison between image-level classification (ILC) accuracy of detections from the common pipeline (\textcolor{ForestGreen}{green}), MIDN (\textcolor{blue}{blue}) and SCC algorithm  (\textcolor{red}{red}) under different training iterations.} 
	\label{img:miscls}
    \vspace{-0.1in}
 
\end{figure}

\subsection{Self-Classification Correction}
A common pipeline for WSOD inference involves two main steps: 1) aggregating the online MCC scores (\textit{e.g.}, OICs \& classification scores in R-CNN branch) as the final results; 2) refining the boxes through the regression outputs. In contrast, the MIDN result is empirically discarded due to its inadequate detection performance. We often observe a common flaw in such a solution:  Some non-exist categories are incorrectly assigned high scores in the final results, leading to many mis-classification samples. We conduct a toy experiment to empirically show the suboptimal classification of final results. We perform statistical analysis on the occurrences of all detected boxes' predicted categories within the image. In other words, if the category appears in the image, the box will be considered a positive sample; otherwise, it is deemed a negative sample. This methodology allows us to  calculate the image-level classification (ILC) accuracy of the bounding boxes. We evaluate multiple models at different training iterations, as shown in Figure \ref{img:miscls}, and compare the detection results from the conventional pipeline (green line) with those from MIDN (blue line).

\begin{algorithm}[t]
\caption {Self-Classification Correction (SCC) }
$\mathcal{C} $ is the set of categories. $\mathcal{O} =  \{O_1, ..,O_K\}$ is the list of online multi-class classification scores. $\mathcal{M}$ is MIDN results.  $\mathcal{F}$ is the final results.   $\lambda$ is the scoring factor. $\tau_{midn}$ is the scoring threshold for MIDN. $O_k, \ \mathcal{F} \in \mathbb{R}^{N\times (|\mathcal{C}|+1)}, \mathcal{M} \in \mathbb{R}^{N\times |\mathcal{C}|}$.
The lines in  {\color{ForestGreen}green} are SCC.

\label{alg:SCC}
\begin{algorithmic}
\State \textcolor{ForestGreen}{$IND \leftarrow$ $\left\{i| \max_c \mathcal{M}_{i,c} > \tau_{midn} \right\} $} \Comment{confident proposals}
\State \textcolor{ForestGreen}{$C_P \leftarrow$ $\arg\max_c$ $\mathcal{M}_{IND,c}$} \Comment{instance-level classification}
\State \textcolor{ForestGreen}{ $C_I \leftarrow$ unique $C_P$}  \Comment{image-level classification}
\State \textcolor{ForestGreen}{ $C_N \leftarrow \mathcal{C} - C_I$}  \Comment{non-exist categories}

\State $\mathcal{F} = mean(\mathcal{O})$

\For {$c$ in range ($C_N$)}
\State \textcolor{ForestGreen}{$\mathcal{F}_{:, c} \leftarrow \lambda\mathcal{F}_{:, c}$}  \Comment{reduce scores}
\EndFor
\State\textbf{return} {$\mathcal{F}$}
\end{algorithmic}
\end{algorithm}

Intrigued, two observations stand out: 1) the ILC accuracy of pipeline result falls short of the 50\% mark, and 2) the MIDN result exhibits notably higher  ILC accuracy compared to the pipeline one. We ascribe this disparity to two main factors. On one hand, only parts of confident proposals participate in the training of the OICs, thus the penalty for non-existent categories has been attenuated. On the other hand, MIDN employs binary cross-entropy loss for the summation of the instance-level scores, which exerts stronger constraints on the image-level classification.

Based on this observation, we propose a simple yet effective strategy to refine the pipeline results with MIDN, termed as Self-Classification Correction (SCC). 
Briefly, we first predict the existing image-level categories according to MIDN results through confident proposal selection and an \text{argmax} operation, and then obtain the absent categories $C_N$.
Subsequently, we reduce the pipeline scores of these absent categories by multiplying a scoring factor $\lambda$.
The details are shown in Algorithm \ref{alg:SCC}. 
As shown in Figure \ref{img:miscls}, SCC algorithm (red line) brings about substantial enhancements in ILC accuracy compared to the pipeline (green line), showcasing improvements ranging from 20\% to 35\%. Remarkably, pipeline utilizing SCC  even achieves superior ILC accuracy when compared to MIDN. Consequently, this refinement strategy contributes to a notable reduction in instances of mis-classification, thus enhancing detector performance without introducing additional training parameters.

Additionally, to make the SCC algorithm more effective, we adopt the misclassification tolerance (MCT) strategy \cite{wu2024misclassification} to further improve the classification ability of MIDN. The motivation behind the MCT strategy is to introduce tolerance for unrepresentative samples with high semantic similarity to an incorrect class, thereby preventing these samples from dominating the training process and forcing the model to memorize them. Suppose $N_p$ classes are present in the image, we identify the misclassified classes containing unrepresentative samples when their score rankings fall within the range of $[N_p, N_p+T_n]$, and assign their class weights to $a$ when calculating $L_{MIDN}$. Similarly, the corresponding incorrect classes, whose score rankings fall within the range of $[0, N_p]$, are also assigned the same class weight.

\section{Experiments and Analysis}

\subsection{Datasets}
Following previous works, we evaluate our proposed method on two popular object detection datasets Pascal VOC 2007 and Pascal VOC 2012 \cite{voc2007}, which contain 20 categories. For both two datasets, we train on \textit{trainval} splits (5,011 images in VOC 2007 and 11,540 images in VOC 2012) and applies two kinds of metrics for evaluation: (1) The mean of average precision (mAP) on the \textit{test} split (4,951 images in VOC 2007 and 10,991 images in VOC 2012); 2) Correct localization (CorLoc) on the \textit{trainval} split. Only image-level labels are utilized during training.

\begin{table}[t!]
        \small
        \vspace{-0.3em}
	\centering
        \renewcommand{\arraystretch}{1.2}
	\setlength{\tabcolsep}{5.0pt}{
	{
			\begin{tabular}{l | c | c}
                \specialrule{.15em}{.05em}{.05em}
				\hline
				Methods & VOC 2007 & VOC 2012 \\ \hline
				OICR~\cite{OICR} & 41.2  & 37.9 \\
				WS-JDS \cite{WS-JDS} & 45.6  & 39.1\\
				C-MIL~\cite{c-mil} & 50.5 & 46.7 \\
				Yang \textit{et al.}  \cite{Yang_2019_ICCV} & 51.5 & 46.8 \\
				C-MIDN~\cite{C-midn} & 52.6 & 50.2 \\
				Pred Net \cite{PredNet} & 52.9 & 48.4 \\
				SLV \cite{SLV} & 53.5 & 49.2 \\
				WSOD$^2$~\cite{WSOD2} & 53.6 & 47.2 \\
				CASD~\cite{CASD} & 56.8 & 53.6 \\
				MIST~\cite{MIST} & 54.9 & 52.1   \\
				IM-CFB \cite{IMCFB}  & 54.3 & 49.4 \\
				SPE~\cite{liao2022end} & 51.0 & - \\
				ODCL~\cite{seo2022object} & 56.1 & \bm{\textcolor{blue}{54.6}} \\
				CBL~\cite{CBL} & \bm{\textcolor{blue}{57.4}} & 53.5 \\
				NDI-MIL~\cite{wang2024negative} & 56.8 & 53.9 \\
				\hline
				\textbf{Ours} & \bm{\textcolor{red}{58.2}} & \bm{\textcolor{red}{55.5}}  \\
				\specialrule{.15em}{.05em}{.05em}
		\end{tabular}}
	}
        \vspace{-0.05in}
        \caption{Performance comparison among the state-of-the-art methods on PASCAL VOC 2007 and  2012. These models are evaluated in terms of mAP (\%). We highlight the best and second best performance in the \textcolor{red}{red} and \textcolor{blue}{blue} colors.}
	\vspace{-0.1in}
	
	\label{tab:map}
	
\end{table}

\subsection{Implementation Details}
Following a widely-used setting, we adopt VGG16 \cite{vgg} pre-trained on ImageNet \cite{imagenet} as the backbone and Selective Search \cite{SS} for proposal generation. The whole framework is end-to-end optimized using stochastic gradient descent (SGD), and the momentum, weight decay, and batch size are set as 0.9, $5 \times 10^{-4}$, and 4, respectively. The initial learning rate is set as $1 \times 10^{-3}$ for the first 70$k$, 170$k$ iterations, and it is dropped by a factor of 10 for the following 20$k$, 40$k$ iterations for VOC 2007 and VOC 2012, respectively. 
We set $\alpha$  to 0.9 and the NMS threshold $\tau_{nms}$ to 0.1 in the SEC module. $\lambda$ and $\tau_{midn}$ in SCC algorithm are set to 0.01 and 0.001, respectively. The hyperparameters of MCT strategy are set as the same with \cite{wu2024misclassification}, \textit{i.e.}, $T_n=1, a=0.4$.
The loss weight $\gamma$ is set to 0.1 for the training balance. Following the previous WSOD works, $\tau_l$, $\tau_h$, and $T$ are set to 0.1, 0.5, and 3, and the images are multi-scaled to \{480, 576, 688, 864, 1000, 1200\} for both training and
inference. 

\begin{table}[t!]
    \small
	\centering
	\renewcommand{\arraystretch}{1.2}
	\setlength{\tabcolsep}{5.0pt}{
	{
			\begin{tabular}{l | c | c}
                \specialrule{.15em}{.05em}{.05em}
				\hline
				Methods & VOC 2007 & VOC 2012 \\ \hline
				OICR~\cite{OICR} & 60.6 & 52.1 \\
				C-MIL~\cite{c-mil} & 65.0  & 67.4 \\
				Yang \textit{et al.}  \cite{Yang_2019_ICCV} & 68.0 & 69.5 \\
				C-MIDN~\cite{C-midn} & 68.7 & 71.2 \\
				WSOD$^2$~\cite{WSOD2} & 69.5 & 71.9 \\
    		SLV \cite{SLV} & 71.0 & 69.2 \\
				MIST~\cite{MIST} & 68.8 & 70.9 \\
                CASD ~\cite{CASD} & 70.4 & 72.3 \\
				IM-CFB \cite{IMCFB}  & 70.7 & 69.6\\
				SPE~\cite{liao2022end} & 70.4 & - \\
				ODCL~\cite{seo2022object} & 69.8 & 71.2 \\
				CBL~\cite{CBL} & \bm{\textcolor{blue}{71.8}} & \bm{\textcolor{blue}{72.6}} \\
				NDI-MIL~\cite{wang2024negative} & 71.0 & 72.2 \\
				\hline
                \textbf{Ours} & \bm{\textcolor{red}{71.9}} & \bm{\textcolor{red}{73.4}}\\
				\specialrule{.15em}{.05em}{.05em}
		\end{tabular}}
	}
        \vspace{-0.05in}
        \caption{Performance comparison among the state-of-the-art methods on PASCAL VOC 2007 and  2012. These models are evaluated in terms of CorLoc (\%). We highlight the best and second best performance in the \textcolor{red}{red} and \textcolor{blue}{blue} colors.}
    \vspace{-0.1in}
	\label{tab:corloc}
\end{table}

\begin{figure*}[t!]
	\centering
    \includegraphics[width=1.0\textwidth,height=0.3\textwidth]{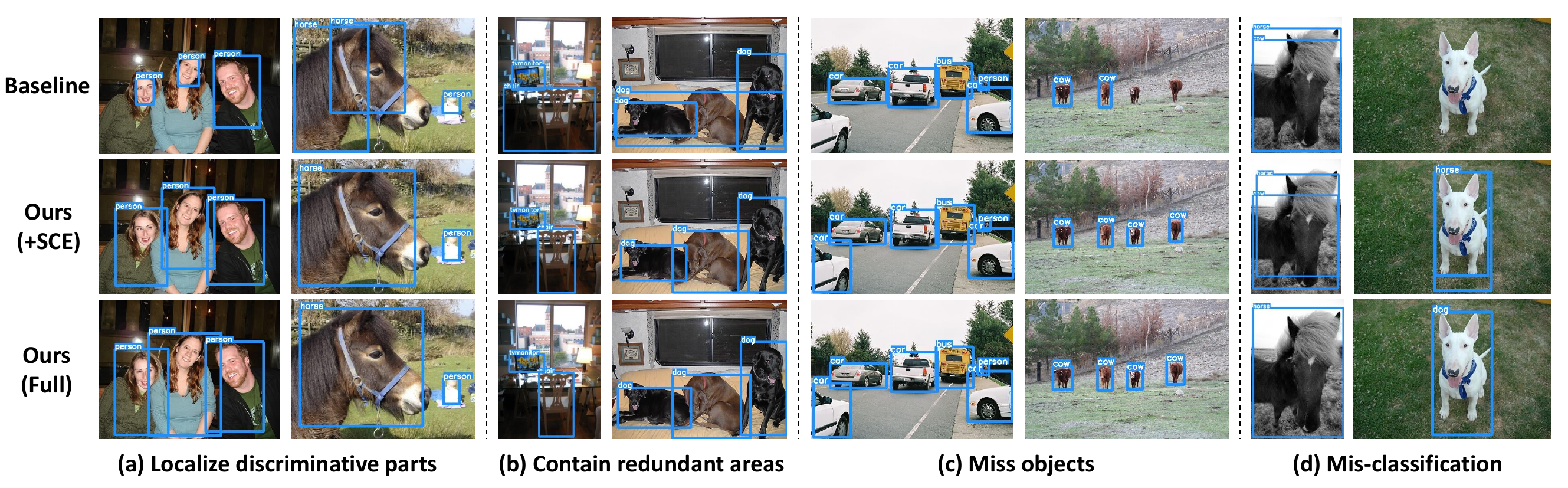} \\%
	\vspace{-0.1in}
	\caption{Qualitative results of the baseline model (1st row), the model only adding SEC module (2nd row), and our whole framework (3rd row) .We show the cases including four typical challenges in WSOD: (a) Localizing only discriminative parts; (b) Containing redundant areas; (c) Missing objects; (d) Mis-classification. } 
	\vspace{-0.1in}
	\label{img:vis_cmp}
	\setlength{\belowcaptionskip}{-6.5pt} 
\end{figure*}

\begin{figure*}[t!]
	\centering
    \includegraphics[width=0.99\textwidth,height=0.2\textwidth]{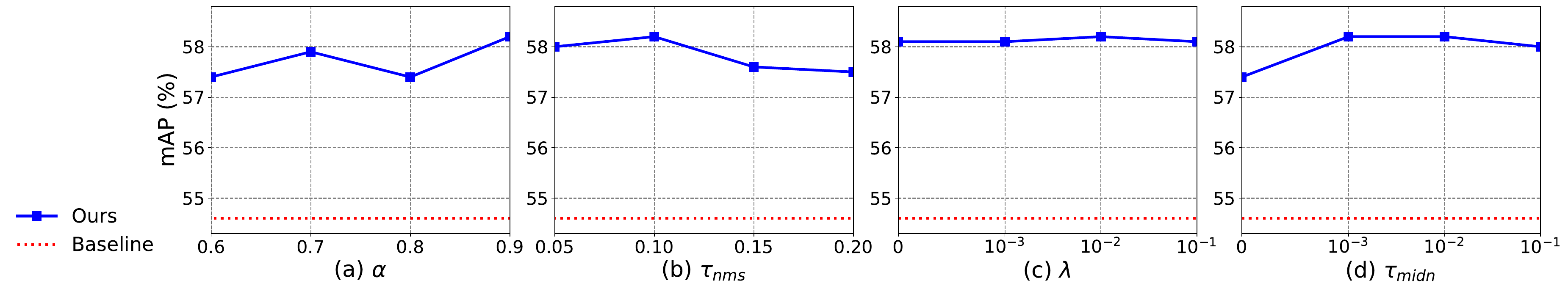} \\%
	\vspace{-0.1in}
	\caption{Visualization results on VOC 2007 \textit{test} set. Successful predictions and failure cases are colored in yellow and green, respectively.} 
	\vspace{-0.1in}
	\label{img:igsm}
	\setlength{\belowcaptionskip}{-6.5pt} 
\end{figure*}

\subsection{Comparison with State-of-the-art Methods}
In Table \ref{tab:map}, we compare the performance of the state-of-art methods with single model on Pascal VOC 2007 and VOC 2012 datasets in terms of mAP. Our method achieves state-of-the-art performance with 58.2\% mAP on VOC 2007 and 55.5\% mAP on VOC 2012, surpassing other methods by at least 0.8\% and 0.9\%, respectively. 
Our method also achieves outstanding performance in terms of CorLoc, setting new state-of-the-art benchmarks with 71.9\% on VOC 2007 and 73.4\% on VOC 2012.
Our method outperforms recent works  \cite{MIST} and \cite{SLV} that improve the seed-box mining but remain reliant on the multi-class classification scores. On one hand, our method considers the scoring variance among different categories and images, thus making a reasonable balance between the accuracy and recall of seed boxes. On the other hand, we use the ICBC scores for further refinement, thus improving the localization accuracy of these seed boxes. Some other methods \cite{OIM,WSOD2} also apply other information for assistance to improve the seed-box mining, however, their improvement limits on the online training procedure. Instead, our  method brings  intra-class binary classification (ICBC) task to directly enhances the network’s discrimination between intra-class positive and negative samples, thus benefiting the feature representation of the whole network.

\begin{table}[t]
    \centering
    \footnotesize
    \renewcommand{\arraystretch}{1.3}
    	\setlength{\tabcolsep}{3.10mm}{
        {
            \begin{tabular}{l  l}
            \specialrule{.15em}{.05em}{.05em}
            Method & mAP (\%)\\
            \hline
            Baseline & $54.6$ \\
            \hline
            \multicolumn{2}{l}{\textit{Self-Classification Enhancement }} \\
            + Intra-class binary classification & $55.2_{\uparrow 0.6}$\\
            + ICBC guided seed mining & $56.7_{\uparrow 1.5}$\\
            \hline
            \multicolumn{2}{l}{\textit{Self-Classification Correction }} \\
            + Self-classification correction & $57.5_{\uparrow 0.8}$\\
            + Misclassification tolerance & $58.2_{\uparrow 0.7}$ \\
            \specialrule{.15em}{.05em}{.05em}
            \end{tabular}
        }
        }
    \vspace{-0.02in}
    \caption{Ablation study of different components of our method on VOC 2007 in terms of mAP (\%).}
    \label{tab:abl}
    \vspace{-0.1in}
\end{table}

\subsection{Ablation Study}
{\flushleft \bf Effect of Each Component.}
We present experimental results on VOC 2007 to validate the effectiveness of each component, as summarized in Table \ref{tab:abl}. Starting with the basic WSOD framework, referred to as the ``baseline", we achieve an initial mAP of 54.6\%. Incorporating the ICBC task into the WSOD framework leads to a notable improvement of 0.6\%, underscoring the effectiveness of the ICBC branches in enhancing the network's ability to distinguish between positive and mislocated samples in a class-wise manner. Subsequently, integrating the IGSM algorithm further improves seed box quality, boosting performance to an mAP of 56.7\%. Overall, the self-classification enhancement module delivers a significant mAP gain of 2.1\%.

To evaluate the self-classification correction module, we first apply the self-classification correction algorithm during inference, resulting in a clear 0.8\% mAP improvement. This gain can be attributed to the algorithm’s capability to effectively reduce high-scoring misclassified samples. By introducing the misclassification tolerance strategy, we achieve the highest performance of 58.2\% mAP. These results demonstrate that enhancing the classification performance of MIDN can further expand the potential upper limit of the self-classification correction algorithm.

\begin{table}[t]
    \centering
    \small
    \renewcommand{\arraystretch}{1.3}
    	\setlength{\tabcolsep}{3.10mm}{
        {
            \begin{tabular}{l  l}
            \specialrule{.15em}{.05em}{.05em}
            Method & mAP (\%)\\
            \hline
            \multicolumn{2}{l}{\textit{Intra-class binary classification}} \\
            only with IoU sampling  & $57.3_{\downarrow 0.9}$ \\
            with gridding sampling ($n$=2)  & $58.2_{\ 0.0}$\\
            with gridding sampling ($n$=3)  & $57.9_{\downarrow 0.3}$\\
            \hline
            \multicolumn{2}{l}{\textit{ICBC guided seed mining}} \\
            without scoring strategy  & $57.2_{\downarrow 1.0}$ \\
            without ICBC-guided finetuning & $57.0_{\downarrow 1.2}$\\
            \specialrule{.15em}{.05em}{.05em}
            \end{tabular}
        }
        }
    \vspace{-0.05in}
    \caption{Ablation study of self-classification enhancement module for the ICBC task on VOC 2007 in terms of mAP (\%).}
    \label{tab:sce}
    \vspace{-0.1in}
\end{table}

\begin{figure*}[t!]
	\centering
    \includegraphics[width=0.99\textwidth,height=0.7\textwidth]{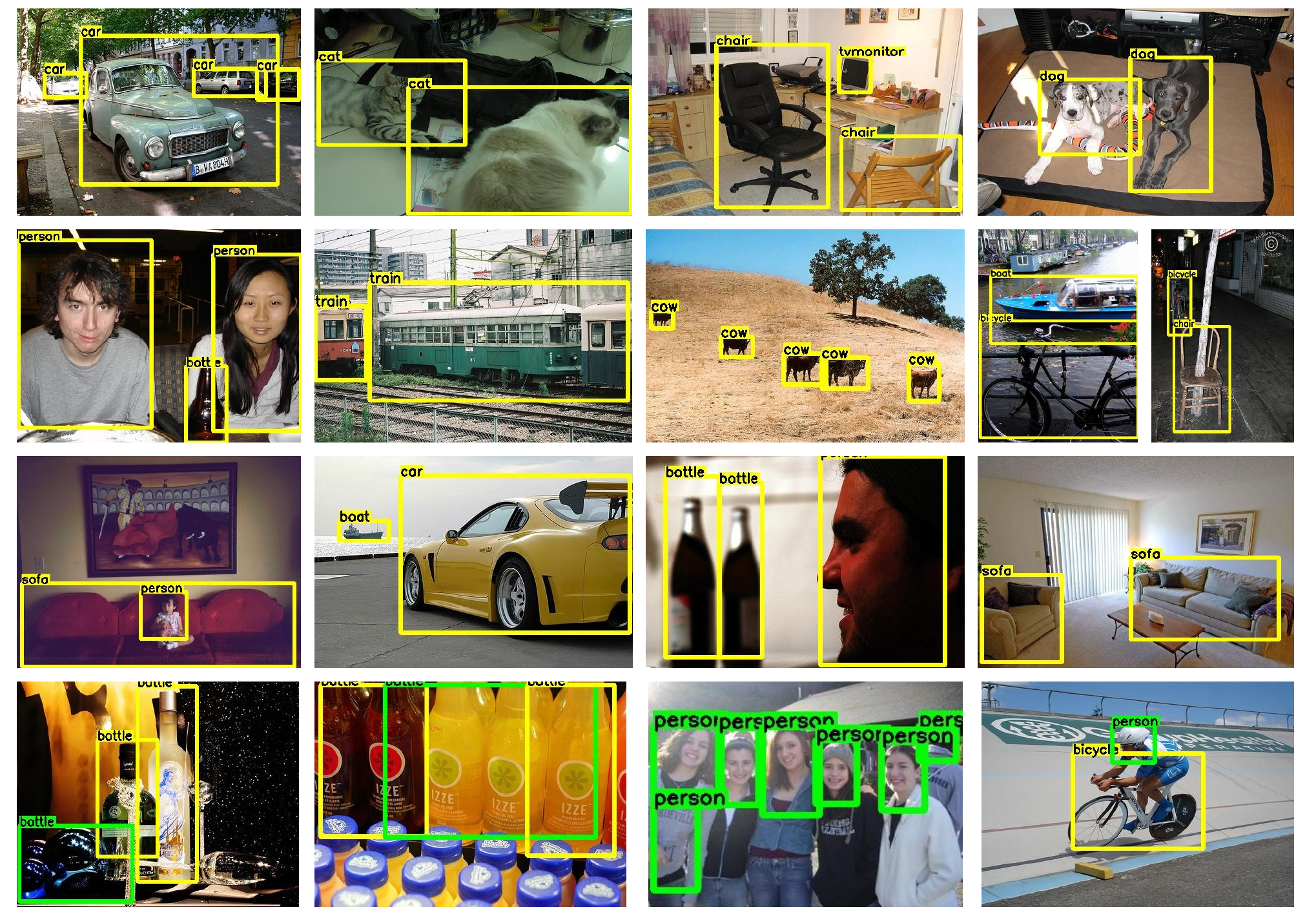} \\%
	\caption{Visualization results on VOC 2007 \textit{test} set. Successful predictions and failure cases are colored in yellow and green, respectively.} 
	\vspace{-0.15in}
	\label{img:vis_test}
	\setlength{\belowcaptionskip}{-6.5pt} 
\end{figure*}

Figure \ref{img:vis_cmp} shows the detection results of different models on VOC 2007 \textit{test} set. Compared with the baseline model (1st row), the integration of  the OEIC module (2nd row) largely alleviates the mis-location problem in different cases, including localizing only discriminative parts (a) and containing redundant areas (including background and other objects) (b). Furthermore, more missing objects are detected (c) due to  the design of the IGSM algorithm in SEC module. Lastly, as shown in (d), the utilization of the MSCR algorithm can further alleviate the mis-classification problem.

{\flushleft \bf Effect of ICBC sampling strategy.}
We conduct experiments by employing different sampling strategies for the training of ICBC task, as shown in the upper part of the Table \ref{tab:sce}. Among all the settings, the combination of the IoU sampling strategy and the gridding strategy achieves the best performance and is insensitive to variations in grid size.

{\flushleft \bf Effect of IGSM algorithm.}
We conduct experiments to evaluate each component of IGSM algorithm, as shown in the lower part of the Table \ref{tab:sce}.  On one hand, we first replace our scoring strategy with original top-$1$ strategy, which only selects the proposals with highest scores as seed boxes, resulting in a 1.0\% mAP drop, as our scoring strategy more effectively locates various objects using class-wise soft thresholds. Additionally, removing ICBC-guided fine-tuning causes a 1.2\% mAP drop, highlighting the superior capability of ICBC in distinguishing positive samples from closely related but mislocated ones.

{\flushleft \bf Effect of Thresholds in IGSM algorithm.}
The first two images in Figure \ref{img:igsm} illustrate the impact of the scoring threshold $\alpha$ and NMS threshold $\tau_{nms}$ in ICBC guided seed mining (IGSM) algorithm. If the restrictions are wide with low $\alpha$ or high $\tau_{nms}$, more noisy samples will be selected as seed boxes, thus exerting negative impacts on the quality of pseudo labels; if the restrictions are tight with high $\alpha$ or low $\tau_{nms}$, the recall of seed boxes will be reduced, which hinders the improvement brought from the IGSM algorithm. 
Compared with the baseline, our IGSM is not sensitive to the choice of values around the optimal ones ($\alpha = 0.9, \tau_{nms}=0.1$), and consistently delivers at least 2.8\% mAP.

{\flushleft \bf Effect of Scoring Factor in SCC algorithm.}
The last two images in Figure \ref{img:igsm} illustrate the impact of the scoring factor $\lambda$ and the scoring threshold $\tau_{midn}$ in the Self-Classification Correction (SCC) algorithm. 
SCC algorithm effectively reduces the mis-classified samples by predicting the non-exist categories and reducing their scores, thus improving the detection performance. 
In general, the performance is not sensitive to the choice of values around the optimal ones ($\lambda = 0.01, \tau_{midn}=0.001$) when the $\tau_{midn}$ is effective ($\tau_{midn}>0$), with the highest gap no more than 0.2\% mAP.
We attribute it to that, SCC algorithm effectively reduces the mis-classified samples by predicting the non-exist categories and reducing their scores, thus improving the detection performance. 

\subsection{Visualization Results}

Figure \ref{img:vis_test} shows the detection results on VOC 2007 \textit{test} set. The first three rows show our successful predictions, which indicates that our method can accurately detect multiple objects in different classes, even if these objects are in some complex backgrounds. The last row shows the failure cases, including localizing only the discriminative parts and grouping different objects. These failure cases especially occur in commonly hard-detected classes, \textit{e.g.}, person and bottle.

\section{Conclusion}
We propose a novel framework for weakly supervised object detection to address the limitations of the two-stage multi-class classification pipeline.  On one hand, we introduce a self-classification enhancement module that enhances the network’s discrimination between intra-class positive and mis-located samples, and leverage it to enhance the quality of seed boxes. On the other hand, we introduce a  self-classification correction algorithm to fine-tune the online classification scores, significantly reducing mis-classification detections. Extensive experiments on the widely used VOC datasets demonstrate the effectiveness of our framework.

\section*{Acknowledgements}
This work was supported in part by the National Natural Science Foundation of China under Grants (62422204, 62472139), in part by the Fundamental Research Funds for the Provincial 
Universities of Zhejiang under Grant No. GK259909299001-040, and in part by the Zhejiang Provincial Natural Science Foundation of China under Grant LQN25F030014, LDT23F02025F02. This work is also supported by the Open Project Program of the State Key Laboratory of CAD\&CG (Grant No. A2403), Zhejiang University.

\bibliographystyle{named}
\bibliography{ijcai25}

\end{document}